\documentclass[]{fairmeta}

\usepackage{amsmath,amsfonts,bm}

\def\eqref#1{equation~\ref{#1}}

\def\1{\bm{1}}

\DeclareMathAlphabet{\mathsfit}{\encodingdefault}{\sfdefault}{m}{sl}
\SetMathAlphabet{\mathsfit}{bold}{\encodingdefault}{\sfdefault}{bx}{n}

\usepackage{url}
\usepackage{booktabs}
\usepackage{tabularx}
\usepackage{float}
\usepackage{graphicx}
\usepackage{listings}
\usepackage{hyperref}
\usepackage{placeins}
\usepackage{xspace}

\lstdefinestyle{prompt}{
  basicstyle=\ttfamily\scriptsize,
  breaklines=true,
  breakatwhitespace=false,
  columns=fullflexible,
  keepspaces=true,
  showstringspaces=false,
  frame=single,
  framesep=4pt,
  aboveskip=0.5em,
  belowskip=0.8em
}

\newcommand{\memorysystem}{memory system\xspace}

\newcommand{\memorysystems}{memory systems\xspace}

\newcommand{\opread}{\ensuremath{\mathtt{read}}}

\newcommand{\opwrite}{\ensuremath{\mathtt{update}}}

\newcommand{\runin}[1]{\noindent{\textbf{#1}}}

\title{Routed Continual Learning}

\author[1,*]{Michael Philip}
\author[1]{Mahesh Balakrishnan}
\author[1]{David Geraghty}
\author[1]{David Mao}
\author[1]{Vidhya Venkat}
\author[1]{Ilya Mironov}

\affiliation[1]{FAIR at Meta}

\contribution[*]{Main Contributor}

\abstract{Agents deployed at scale repeatedly encounter related problems, and it is natural to let them learn from each other's experience rather than starting over each time. Unfortunately, there is no single way to retain that experience that works well across settings, and no way to know in advance which future problems a given experience will help with. A global history preserves every detail but becomes expensive and mixes together experience that may not belong together; compressing that history into one fixed representation limits the cost but can throw away valuable details.

We introduce Routed Continual Learning (Routed CL), an architecture that decides how to partition and retain experience. Past trajectories are grouped into clusters, each of which maintains state using one memory system from a fixed library. Before each run, an LLM-based router either reuses an existing cluster or creates a new one, basing its decision on the problem statement, memory system descriptions, and cluster summaries. We also introduce Context Memory Adapter, which mediates between the memory system and the agentic loop by augmenting the agent's initial context with the memory system's outputs and updating the selected cluster with information from the trajectory.

We evaluate the architecture on a heterogeneous stream that interleaves 301 instances from the six CL-Bench tasks with 97 banking instances from $\tau^3$-Bench, for 398 instances total. Across four language models and five matched rollouts, Routed Continual Learning raises normalized reward over a single global in-context learning (ICL) history by 15--30\% of the maximum attainable normalized reward, achieves higher mean normalized reward than ICL-only clustering, and reduces estimated API cost by 65--72\% relative to Global ICL. These results support approaching continual learning as a routing problem that chooses both a learning mechanism and a state cluster rather than committing to one representation globally.
}

\date{October 2026}
\correspondence{Michael Philip at \email{michaelphilip@meta.com}}

\begin{document}

\maketitle

\section{Introduction}

State-of-the-art agentic systems are designed to operate at scale, with 
distributed heterogeneous agents acting concurrently and independently. Many of these agents
encounter related problems and could benefit from shared knowledge, which can
reduce token cost, improve task performance, and keep agents from repeating one
another's mistakes. One striking example occurred during a recent OpenAI cyber-capability evaluation. Independently instantiated agents created an unintended shared message board and used it to propagate exploit findings across runs. The resulting exchange of information enabled agents in the broader evaluation to obtain breakthrough capabilities beyond the reach of individual agents~\citep{openai2026huggingface,greenblatt2026huggingface}.

Our goal is to make such sharing deliberate: to build a system that supplies agents with relevant prior experience by design, rather than expecting them to develop one. To this end, we must decide what information to extract from trajectories, how to represent it, which agents should receive it, and when it should be provided. Different applications may require different answers to these questions; no single \memorysystem is optimal across all of them. For example, coding-heavy tasks may benefit from a repo index, tool-oriented workloads from examples, knowledge tasks from grounded facts, data analytics from time-series models, etc.

The distributed environment of an enterprise setting compounds these challenges: agents are built by different teams, wrap different models, and have no reason to know that other agents exist. Our architecture places minimal requirements on the agents it supports. It intercepts an agent's prompt and, when useful, injects relevant content directly into the agent's context, requiring no explicit cooperation.

The result of this work is an architecture with two central features:

\begin{itemize}
\item \textbf{Transparent deployment:} The architecture requires neither coordination among individual agents nor changes to how they operate.
\item \textbf{Routing:} Because no single memory system serves every application equally well, a routing component selects a memory system and relevant prior state for each run.
\end{itemize}

\begin{table}[t]
  \centering
  \footnotesize
  \setlength{\tabcolsep}{3pt}
  \renewcommand{\arraystretch}{1.16}
  \caption{Normalized reward across five matched rollouts. Each
  cell reports the mean and the range as $[\min,\max]$. Stateless-reference
  gain is reported separately in Table~\ref{tab:fixed-reference-gain}. Higher
  is better; bold marks the best of the four staged conditions. Ground-truth
  assignment uses source-task identity, unavailable to the router, to apply a
  fixed memory-system--cluster mapping specified before confirmatory evaluation.}
  \label{tab:main-results}
  \begin{tabularx}{\linewidth}{@{}l*{4}{>{\centering\arraybackslash}X}@{\hspace{5pt}\vrule width 0.35pt\hspace{5pt}}>{\centering\arraybackslash}X@{}}
    \toprule
    Model & Stateless ICL & Global ICL & Routed ICL & Routed CL & \shortstack{Ground-truth\\assignment} \\
    \midrule
    \shortstack{Gemini 3.7 Flash \\{\scriptsize\phantom{$[0,0]$}}}
& \shortstack{$-0.055$\\{\scriptsize$[-0.066,-0.041]$}} & \shortstack{$-0.030$\\{\scriptsize$[-0.055,-0.022]$}} & \shortstack{0.211\\{\scriptsize$[0.199,0.245]$}} & \shortstack{\textbf{0.213}\\{\scriptsize$[0.111,0.272]$}} & \shortstack{0.284\\{\scriptsize$[0.236,0.333]$}} \\
\shortstack{GPT-5.6 Sol  \\{\scriptsize\phantom{$[0,0]$}}}
    & \shortstack{0.130\\{\scriptsize$[0.122,0.137]$}} & \shortstack{0.168\\{\scriptsize$[0.125,0.189]$}} & \shortstack{0.387\\{\scriptsize$[0.333,0.424]$}} & \shortstack{\textbf{0.464}\\{\scriptsize$[0.442,0.501]$}} & \shortstack{0.478\\{\scriptsize$[0.441,0.531]$}} \\
\shortstack{Opus 4.7  \\{\scriptsize\phantom{$[0,0]$}}} & \shortstack{0.060\\{\scriptsize$[0.051,0.067]$}} & \shortstack{0.092\\{\scriptsize$[0.077,0.104]$}} & \shortstack{0.209\\{\scriptsize$[0.189,0.242]$}} & \shortstack{\textbf{0.243}\\{\scriptsize$[0.184,0.309]$}} & \shortstack{0.312\\{\scriptsize$[0.276,0.370]$}} \\
\shortstack{Opus 5 \\{\scriptsize\phantom{$[0,0]$}}} & \shortstack{0.178\\{\scriptsize$[0.155,0.196]$}} & \shortstack{0.169\\{\scriptsize$[0.152,0.198]$}} & \shortstack{0.340\\{\scriptsize$[0.302,0.383]$}} & \shortstack{\textbf{0.401}\\{\scriptsize$[0.356,0.433]$}} & \shortstack{0.408\\{\scriptsize$[0.392,0.423]$}} \\
    \bottomrule
  \end{tabularx}
\end{table}

Continual Learning Bench (CL-Bench;~\citealp{asawa2026clbench}) provides a useful starting point for evaluating an implementation of this architecture. Following CL-Bench's convention, we call each individual problem an \emph{instance}; a sequence of instances from the same domain is a \emph{task}. Examples of these tasks include text-to-SQL transformations and playing poker against deterministic opponents. An instance of a task for the first example is a single natural-language question that may require multiple SQL queries to resolve; an instance of the poker task is a single hand. CL-Bench evaluated six such tasks in isolation and found that na\"ive in-context learning (ICL) outperformed dedicated memory systems on the aggregate metric. Retaining prior trajectories provides a strong baseline when each task is evaluated separately, which is the target scenario for the benchmark. This approach becomes less effective when different tasks are interleaved: a single history combines experience that may be useful for one task but irrelevant for another. A learning system must therefore decide what observations to share among which instances, and when to use a different \memorysystem.

To expand the range of tasks beyond CL-Bench, we incorporate the Banking Knowledge domain from $\tau^3$-Bench~\citep{shi2026tauknowledge}. Each instance is a multi-turn customer-service conversation in which an agent must follow a domain policy, retrieve unstructured knowledge, and use tools while interacting with a simulated user. This contributes 97 thematically related instances with an interaction structure and reward function substantially different from those of the six CL-Bench tasks.

We introduce \emph{Routed Continual Learning}, which assigns each instance to a \emph{cluster} that couples a \memorysystem with a set of prior trajectories (that can be empty). Before each instance, a \emph{router}, implemented as a large language model (LLM), decides whether to reuse an existing cluster or create a new one, in which case the router selects a \memorysystem from a fixed library. The \memorysystem defines the learning mechanism; the cluster maintains the \memorysystem's state and publishes a summary to guide assignment of future instances. The router uses only the current instance, \memorysystem descriptions, and cluster summaries; task labels during the benchmark evaluation are not supplied.

Our five \memorysystems (Table~\ref{tab:operator-library}) retain prior trajectories, distill reusable reasoning
strategies, or model structured numerical state. Some \memorysystems require mapping a task-specific schema into generic state. We therefore
introduce \emph{Context Memory Adapter}, a shared runtime that derives this mapping, or
adapter, from prior experience and supplies learned state through the agent's
context. 

We evaluate four language models on a heterogeneous 398-instance stream
combining all six CL-Bench tasks with 97 $\tau^3$-Bench banking instances. Four
staged conditions progressively add capabilities: Stateless ICL (no
cross-instance state), Global ICL (one shared history), Routed ICL
(router-partitioned ICL histories), and Routed CL (the full \memorysystem
library). We compare these conditions on normalized reward and API cost.

Our contributions are:
\begin{itemize}
\item We formulate continual learning over heterogeneous, interleaved instance sequences as a joint decision: how to represent experience and which instances should share state.

\item We introduce Routed Continual Learning, which for each instance selects a \memorysystem from a fixed library and a cluster to hold its resulting state.

\item We show that across four language models and five matched rollouts, Routed CL improves normalized reward over both Global ICL and Routed ICL while reducing estimated API cost by 65--72\% relative
to Global ICL.
\end{itemize}

\section{Related Work}

\runin{Analysis of algorithms and learning theory.} Before turning to agentic systems, we recall earlier lines of work that share the same central insight: An algorithm or system tuned for a single-shot task differs from one optimized holistically for a sequence of related tasks, which is arguably the more practically relevant regime. By retaining state across tasks, the system can learn from experience, amortize computational effort, transfer knowledge, and adapt to the empirical task distribution.

The amortized complexity analysis of algorithms and data structures~\citep{tarjan-amortized85} considers the average cost of a sequence of operations rather than their peak complexity, establishing, for instance, competitive guarantees for the LRU caching policy and amortized guarantees for splay trees~\citep{SleatorTarjan85-caching,
SleatorTarjan85-splay}. Self-improving algorithms adapt to the data distribution~\citep{Ailon_etal-self-improving06}, converging to the optimal expected complexity. 

In online learning, samples are revealed incrementally; the main challenge is to perform competitively relative to a baseline chosen in hindsight after observing the entire sequence~\citep{SSS2012-online}. Continual learning concerns models that learn from a sequence of tasks or changing data distributions, aiming to adapt to new information while retaining previously acquired knowledge~\citep{Parisi_etal2019-continual_learning}.

\runin{Prompt rewriting and context augmentation.} Model performance can be improved at inference time by adapting both how a task is specified and what information is supplied to the model. Automatic Prompt Engineering (APE) shows that automatically generating task instructions can outperform manually designed prompts~\citep{zhou2023-APE}; more recently, GEPA demonstrates that prompts can be iteratively improved through LLM-based reflection on execution trajectories~\citep{agrawal2026gepa}. Eevee jointly optimizes a router and a set of prompts, coupling routing to prompt updates so that each prompt specializes to the inputs it receives~\citep{xu2026eevee}. APE and GEPA optimize prompts offline, over a batch of accumulated tasks; Eevee assumes ground-truth labels for training its router. Routed Continual Learning makes online choices, selecting a \memorysystem and a cluster for each instance as it arrives, from a fixed library whose members we do not optimize.

Context augmentation enriches the model's input with retrieved knowledge or sampled outputs~\citep{Ram_etal2023-in-context,luo_etal2024-demonstrations-survey}. Dynamic Cheatsheet maintains a single evolving memory of strategies and solutions carried across problems~\citep{suzgun_etal2026-cheatsheet}. Agentic Context Engineering (ACE) grows this memory through incremental edits rather than full rewrites, which reduces the risk of context collapse that repeated rewriting can cause~\citep{zhang_etal2026-ace}. Both systems accumulate one context across all problems; Routed Continual Learning instead partitions accumulated state into separate clusters and selects among them, so that unrelated problems need not share a history.

\runin{Memory systems.} Systems that give agents access to past experience differ mainly in what they keep: raw context, distilled facts, advice, structured procedures, or latent vectors. Routed Continual Learning is compatible with all of them: it picks a \memorysystem per instance from a fixed library, and keeps each pick's state in its own cluster.

MemGPT treats the context window like physical memory, paging information in and out of external storage as the agent requests it through function calls~\citep{packer2023memgpt}, while CoEvo-Mem learns the retrieval policy itself, adjusting how memories are fetched as it refines what they contain~\citep{ye2026coevomem}. Mem0 pulls relevant facts from each trajectory and reconciles them against what it already holds, adding, revising, or deleting entries as it goes~\citep{chhikara2025mem0}. 

Reflexion writes a short self-critique after a failed attempt and reads it back on the next try~\citep{shinn2023reflexion}; ExpeL abstracts insights that hold across tasks~\citep{zhao2024expel}; Dynamic Cheatsheet keeps a running note of strategies that have worked~\citep{suzgun_etal2026-cheatsheet}; ACE grows a playbook of scored bullets, editing it in small increments so earlier entries survive~\citep{zhang_etal2026-ace}; and ReasoningBank mines both successes and failures for reusable reasoning strategies~\citep{ouyang2026reasoningbank}. \citet{ibm2026-trajectory-memory} propose trajectory-informed memory generation that groups experience by outcomes: clean successes, runs that failed and recovered, and runs that succeeded wastefully.

Other systems retain something other than free-form text from past executions. Agent Workflow Memory extracts reusable workflows~\citep{wang2024awm}; Voyager accumulates a library of executable skills~\citep{wang2023voyager};  MemGen maintains a latent space representation~\citep{zhang2026memgen}.


\runin{Standard harness hooks and rewriters.} Most agent frameworks already expose a point at which a message can be intercepted and modified before an agent's own logic runs, without changing the agent itself. AutoGen's ConversableAgent, for instance, lets a caller register hooks that fire before a reply is generated, including one that updates the agent's state~\citep{wu2024autogen}. Production deployments extend the same idea into guardrail and middleware layers, where an input hook inspects or rewrites a message on its way to the model and an output hook does the same to the response, again without modifying the agent's own code. In our implementation, Context Memory Adapter attaches at exactly this point: rather than validating or filtering a message, it augments the prompt to include retrieved memory content, using the same hook mechanism these frameworks already provide for other purposes.

\section{Problem Formulation}
\label{sec:problem}

Routed continual learning addresses settings in which neither the appropriate representation of past experience nor the prior state relevant to a new instance is known in advance. We consider a causally ordered, heterogeneous stream containing interleaved instances from several source tasks, without explicit task labels. Using only the visible instance specification, \memorysystem descriptions, and cluster summaries, the system must determine how to represent experience and which instances should share state.

\runin{Heterogeneous continual streams.}
Continual Learning Bench (CL-Bench) defines a task as an ordered sequence of
problem instances \citep{asawa2026clbench}. We study a setting in which instances
from multiple such source tasks are interleaved into one heterogeneous stream.
Let source task $d \in \{1,\ldots,D\}$ provide instances
$x_{d,1},\ldots,x_{d,N_d}$. The evaluated stream $x_1,\ldots,x_N$, where
$N=\sum_{d=1}^{D}N_d$, preserves the order of instances within each source task.
At position $t$, the system receives the visible specification of $x_t$: its
natural-language prompt and required structured response format. It does not
receive an explicit task label or information about its relationship
to earlier instances.

Completing an instance produces an agent-visible trajectory $\tau_t$ containing
the instance prompt, model actions, tool outputs, and feedback revealed through
normal interaction. The environment also assigns reward $r_t$ for evaluation,
but does not share it with the learning system. At every decision, the
system may use only information exposed by that point in the interaction.


\runin{Continual-learning \memorysystems.} We call the component of our architecture that retains experience across agentic runs a continual-learning
\memorysystem. Let $\mathcal M$ be a finite library of such systems whose implementations and allowed configuration schemas are fixed before evaluation. A \memorysystem renders text from its state for the actor's context and updates that state by consuming trajectories. Once cluster $c$, backed by system $m$, has been configured, these operations are
\[
z_t = \opread_m(S_{c,t},x_t;\theta_c), \qquad
S_{c,t+1} = \opwrite_m(S_{c,t},\tau_t;\theta_c),
\]
where $S_{c,t}$ is the cluster state, $\theta_c$ is its cluster-specific configuration, and $z_t$ is text added to the actor's context. Only the selected cluster's state is updated. Multiple agents may interact with the same cluster concurrently; \opread\ and \opwrite\ operations are atomic.

Some \memorysystems require task-specific bindings between the task schema and generic state. For those systems, Context Memory Adapter derives the cluster-specific adapter configuration $\theta_c$ from prior runs; other systems use fixed interfaces. Memory systems interact with the agent through context; they neither act in the environment nor modify its parameters.

\runin{Stateful routing.}
The system must choose both a learning mechanism and the cluster whose state the
current trajectory will read and update. A cluster $c$ is an isolated instance
of one \memorysystem, with its own state and an immutable description $\delta_c$
of which instances should share that state; multiple clusters may use the same system. Let $C_t$ be the
clusters available before instance $t$. Given $x_t$, \memorysystem descriptions,
and summaries $\Sigma(C_t)$, the router $\pi$ chooses either
$\mathtt{reuse}(c)$ for $c\in C_t$ or $\mathtt{create}(m,\delta)$ for
$m\in\mathcal{M}$. The choice remains fixed during the instance. New clusters
begin from initial state. Reusing too broadly can mix unrelated state, while
unnecessary creation can prevent useful transfer.

Thus, routing jointly selects a \memorysystem and a cluster: Global ICL uses one
ICL state, Routed ICL selects among ICL clusters, and Routed CL selects both.

\section{System Components}
\label{sec:method}

\begin{figure}[t]
  \centering
  \includegraphics[width=\linewidth]{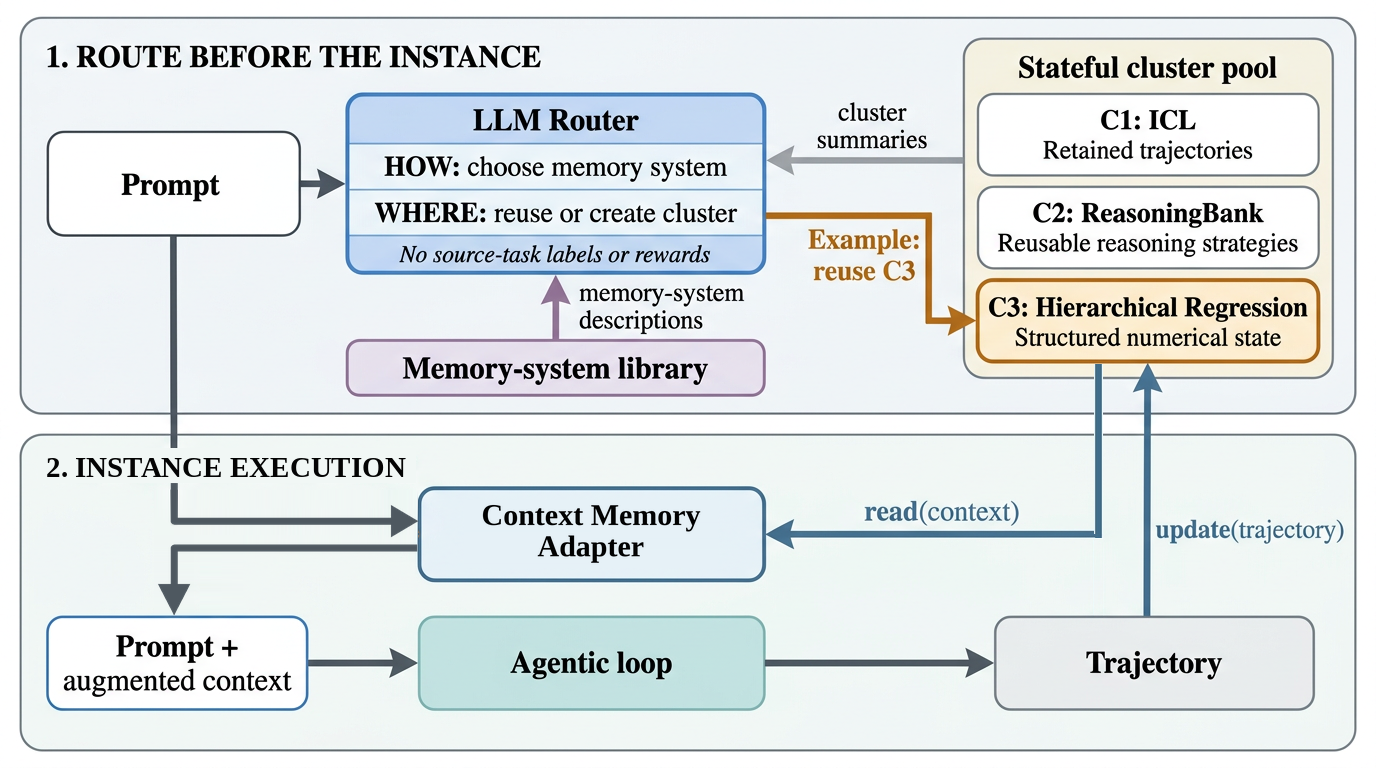}
  \caption{Overview of Routed Continual Learning. Before each instance, the
  router chooses a memory system and either reuses or creates a
  stateful cluster. Context Memory Adapter reads its state and augments the task
  context before execution; the selected cluster is updated from the trajectory,
  while all other clusters remain unchanged. The displayed clusters are
  illustrative; the pool grows as clusters are created and reused online.}
  \label{fig:routed-cl-overview}
\end{figure}

Routed CL comprises an LLM router, isolated
stateful clusters, a fixed memory-system library, and Context Memory Adapter. The
memory system determines how experience is retained, while the cluster determines
which instances share state. Upon receiving an instance, the router chooses an existing cluster or creates a new one; the cluster assignment stays static throughout the instance execution. Figure~\ref{fig:routed-cl-overview} summarizes
this lifecycle.

\subsection{Router and Stateful Clusters}

The router implements $\pi$ with one structured-output LLM call before each
instance. It receives the visible prompt and response schema, each memory
system's declared learning behavior, and, for every cluster, only its identifier,
memory-system type, immutable description, and completed-trajectory count. It
cannot inspect raw trajectories, internal memory-system state, or hidden
conversation state from earlier routing calls.

The router is prompted rather than trained. Its prompt and memory-system library
remain fixed during evaluation, and it receives no source-task labels or rewards.

\subsection{Frozen Memory-System Library}

Mechanism selection is meaningful only when memory systems encode different
assumptions about reusable information. We use five systems spanning three
representation families (Table~\ref{tab:operator-library}). The router chooses
among the five concrete systems; the families only organize the presentation
and do not create another routing layer. These systems are examples rather than
a canonical taxonomy. The library was selected during exploratory development
on CL-Bench and frozen before confirmatory evaluation; our experiments test
selection and configuration within it, not the synthesis of new memory-system
families.

\begin{table}[H]
  \centering
  \small
  \setlength{\tabcolsep}{5pt}
  \renewcommand{\arraystretch}{1.08}
  \caption{Five memory systems grouped by the form of state they retain.}
  \label{tab:operator-library}
  \begin{tabularx}{\linewidth}{@{}>{\raggedright\arraybackslash}p{0.41\linewidth}X@{}}
    \toprule
    Memory system & Retained state and observation structure \\
    \midrule
    \multicolumn{2}{@{}l}{\textbf{\textit{Retained trajectories}}} \\[0.22em]
    \quad In-context Learning (ICL) & Ordered visible trajectories \\
    \addlinespace[0.35em]
    \multicolumn{2}{@{}l}{\textbf{\textit{Distilled reasoning memory}}} \\[0.22em]
    \quad ReasoningBank (RB) & Strategies induced from self-judged successful and failed trajectories \\
    \addlinespace[0.35em]
    \multicolumn{2}{@{}l}{\textbf{\textit{Structured numerical state}}} \\[0.22em]
    \quad Hierarchical Regression (HR) & Repeated keyed scalar outcomes \\
    \quad Hierarchical Discrete-Event State (HDE) & Monotone event-free curves over ordered horizons \\
    \quad Persistent-Entity State (PES) & Noisy measurement sets with uncertain entity correspondence \\
    \bottomrule
  \end{tabularx}
\end{table}

\runin{In-context Learning (ICL).}
Each ICL cluster retains visible prompts, actor responses, and feedback as an
ordered history, discarding the oldest messages only at the model context limit.
Its only abstraction is the router-defined cluster boundary.

\runin{ReasoningBank.}
We adapt ReasoningBank~\citep{ouyang2026reasoningbank} to the Context Memory Adapter
interface. After each completed visible trajectory, the same backbone model
judges whether the attempt succeeded or failed and extracts up to three
structured memory items capturing reusable strategies or failure-avoidance
lessons. Before a later instance, the system retrieves the memories associated
with the most similar prior request and injects them at the instance boundary.
Our adaptation uses no hidden rewards,
uses frozen local embeddings, and does not use ReasoningBank's test-time-scaling
extension.

\runin{Numerical memory systems.}
The numerical systems' statistical libraries consume generic typed records and
contain no CL-Bench task names, task-specific fields, prompts, or scoring logic.
Hierarchical Regression and Hierarchical Discrete-Event State use
PyMC~\citep{abrilpla2023pymc}, and Persistent-Entity State uses River's
Bayesian linear regression~\citep{montiel2021river}; task semantics enter
through adapter bindings and extracted evidence.
Appendix~\ref{app:numerical-operators} provides implementation details.

\subsection{Context Memory Adapter}

Context Memory Adapter configures cluster-specific adapters when a memory system
requires task-specific bindings, and exposes the selected cluster's state
through the actor's context; Appendix~\ref{app:adapter-validation} describes the
adapter lifecycle and validation procedure. An adapter supplies only
schema-bounded semantic bindings and permitted configuration choices, never
parser code or task answers, so online adaptation is limited to routing, bounded
configuration, and state updates within a frozen library.

To isolate continual learning from task execution, Context Memory Adapter never acts
in the environment or submits an answer. Before the first actor turn, it may
inject a budget-bounded advisory alongside the unchanged prompt; the advisory
remains in the within-instance history. The selected cluster is updated only
with information from its assigned trajectory. The same actor retains its original tools and
response schema, and only its response is scored. Context Memory Adapter cannot access
hidden tests or evaluation rewards, so memory affects behavior solely through
context.

\section{Experiments}
\label{sec:experiments}

\runin{Benchmark and heterogeneous stream.} Routed Continual Learning is intended
for streams where prior experience can help, but the system is not told which
memory system to use or which instances should share state. Evaluating the source
tasks separately would test individual learning mechanisms but not the routing
problem. We therefore construct a heterogeneous stream by interleaving
all 301 instances from the six CL-Bench tasks~\citep{asawa2026clbench} with 97
Banking Knowledge instances from $\tau^3$-Bench~\citep{shi2026tauknowledge}, for
398 instances total. CL-Bench contributes six tasks spanning distinct interfaces,
reward functions, and forms of reusable cross-instance information.
$\tau^3$-Bench adds multi-turn customer-service interactions requiring
policy-grounded reasoning, knowledge retrieval, and tool use. The source tasks
are summarized in Table~\ref{tab:clbench-tasks}.

For each rollout, we form one heterogeneous stream by randomly interleaving the seven source-task sequences while preserving each source task's canonical order. All systems are evaluated on the same set of five rollouts. Source-task labels used for analysis are not supplied to the router or agents. Native prompts, tools, response schemas, feedback, and reward functions remain visible, so routing may use observable task structure.

\runin{Compared systems.} We compare four staged conditions with matched actors, task interfaces, budgets,
model limits, and stream order. Stateless ICL resets after
each instance; Global ICL shares one history across the stream; Routed ICL partitions that history among dynamically defined clusters; and Routed CL also selects among the five memory systems in Table~\ref{tab:operator-library}.
Configurable clusters derive their adapters from observable warm-up experience on first reuse. All ICL histories use a 100K-token context budget, applied stream-wide for Global ICL and independently per cluster for the routed conditions.

Separately, we evaluate our ReasoningBank adaptation under two fixed-system configurations. Global RB maintains one memory store across the stream, whereas Routed RB maintains one store per router-selected cluster. This comparison uses three matched rollouts and is reported in Appendix~\ref{app:reasoning-bank-control}.

We also evaluate Ground-truth assignment, a privileged diagnostic that uses source-task identity and a fixed memory-system and cluster mapping chosen during exploratory development and frozen before confirmatory evaluation.

\runin{Models and repetitions.} To assess whether our findings hold across model families, we evaluate four language models: Claude Opus 4.7, Claude Opus 5, GPT-5.6 Sol, and Gemini 3.7 Flash. Within each condition, every model-mediated role uses the evaluated model: the actor and any router, adapter configurator, typed extractor, ReasoningBank judge, or memory writer required by that condition. The fixed $\tau^3$-Bench user simulator and evaluator are benchmark components shared across conditions.

To estimate variability from stream ordering and stochastic model outputs, we
evaluate each model--condition pair over five complete rollouts. Because persistent state couples later instances to earlier
behavior, the complete rollout is the unit of replication. For each language
model, we summarize the five paired differences between methods.

\runin{Evaluation protocol.} We froze all method components before confirmatory evaluation. Matched
conditions receive the same task interface, feedback, budgets, and model limits;
Appendix~\ref{app:evaluation-contract} gives the complete specification.

All task-derived evidence comes from the native task interface. The four staged
conditions receive no source-task identities, privileged assignments, hidden
tests, or held-out rewards; Ground-truth assignment receives only source-task
identity. To prevent cross-instance leakage, we reset Sales Prediction's
writable workspace and expose Codebase Adaptation actors only to base-commit
snapshots, with regression tests reserved for evaluation. These changes apply
to every method but make the affected results not directly comparable to
historical CL-Bench runs
(Appendices~\ref{app:sales-isolation} and~\ref{app:code-isolation}).

Resource totals include every model call.
Appendix~\ref{app:prompts} reproduces the fixed prompts and memory-system
descriptions.

\runin{Metrics.} We report normalized reward as the primary metric. For the six CL-Bench tasks,
the external stateless reference published with CL-Bench maps to 0 and the task
maximum to 1; scores below the reference are negative. $\tau^3$-Bench instead uses its
native $[0,1]$ reward, and all seven source tasks are weighted equally. Our
secondary metric, stateless-reference gain, compares each method with Stateless ICL
using the same language model and matched interleaving: 0 denotes matched
Stateless ICL performance and 1 the task maximum.
Appendix~\ref{app:metric-definitions} provides the complete definitions.

We estimate API cost for every model call using a frozen, dated public list-price
schedule that accounts for token caching.
Appendix~\ref{app:cost-accounting} gives the complete accounting policy.


\section{Results and Analysis}
\label{sec:results}

We demonstrate that Routed Continual Learning sharply improves aggregate reward
over Stateless ICL and globally shared memory baselines, while costing
substantially less than Global ICL. Detailed task- and model-level breakdowns are
deferred to the appendix.

\runin{Global ICL loses its advantage under heterogeneous interleaving.}
CL-Bench found na\"ive ICL strongest when its tasks were evaluated as separate
homogeneous streams~\citep{asawa2026clbench}. In our interleaving, mean
stateless-reference gain rises
from 0.015 for Global ICL to 0.222 for Routed ICL and 0.280 for Routed CL.
Figure~\ref{fig:learning-curves} shows how this pattern develops over
the stream. For Opus 4.7, Global ICL helps early but its gain declines later;
for Opus 5, it remains worse than Stateless ICL throughout. The routed methods
show larger gains later in the stream.

\runin{Clustering helps ICL but does not consistently help ReasoningBank.}
Routed ICL exceeds Global ICL by 0.117--0.241 normalized-reward points across
models. By contrast, routing ReasoningBank produces no consistent reward gain
and raises mean cost by 8--29\% (Figure~\ref{fig:reasoning-bank-tradeoff}). This may be because ICL exposes an
entire cluster history, whereas ReasoningBank already retrieves memories
selectively by request similarity.

\begin{figure}[H]
  \centering
  \includegraphics[width=\linewidth]{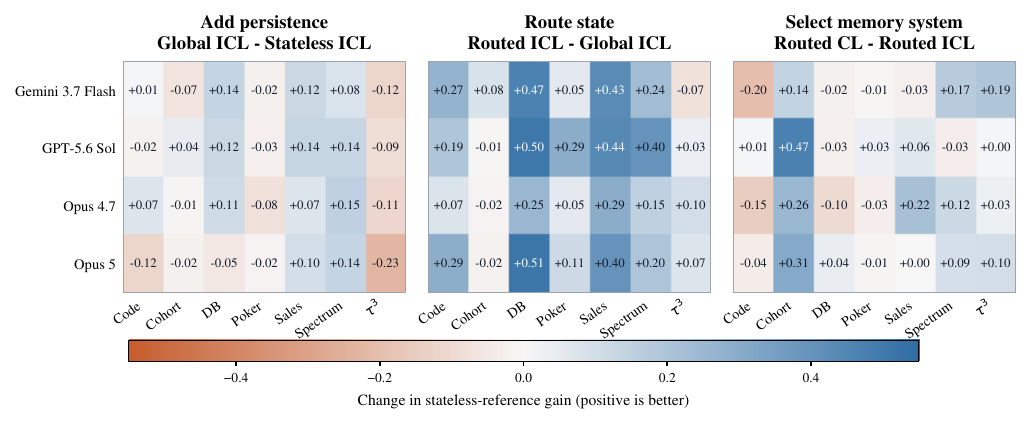}
  \caption{Five-seed mean change in stateless-reference gain by source task. The
  panels compare successive conditions: Global ICL minus Stateless ICL, Routed
  ICL minus Global ICL, and Routed CL minus Routed ICL. Blue indicates
  improvement and orange regression.}
  \label{fig:domain-gain}
\end{figure}

\begin{figure}[H]
  \centering
  \includegraphics[width=\linewidth]{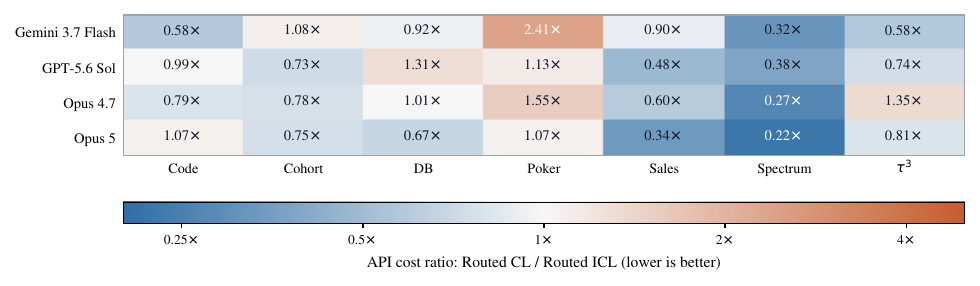}
  \caption{Five-seed mean API-cost ratio between Routed CL and Routed ICL by
  source task. Each cell reports
  $C_{\mathrm{Routed\ CL}}/C_{\mathrm{Routed\ ICL}}$ using mean task-level costs
  across matched rollouts. Values below one (blue) indicate that Routed CL is
  cheaper; values above one (orange) indicate that it is more expensive. Costs
  include all model calls attributed to instances from that source task.}
  \label{fig:task-cost-ratios}
\end{figure}

\runin{Memory-system selection adds task-specific gains.}
Routed CL adds another 0.002--0.076 normalized-reward points. Gemini's mean
improvement is only 0.002 but is positive in four of five paired rollouts; the
other models show larger gains.
Table~\ref{tab:paired-routing-differences} reports the paired results. At the task level, Routed CL improves over Routed ICL on Cohort and
$\tau^3$-Bench for every model and on Spectrum for three of four models, while
reducing cost on Sales and Spectrum for every model and on Cohort for three of
four models (Figures~\ref{fig:domain-gain} and~\ref{fig:task-cost-ratios}). On Sales, for
example, Opus 5 maintains essentially unchanged reward at 0.34$\times$ the~cost.

\begin{figure}[t]
  \centering
  \includegraphics[width=\linewidth]{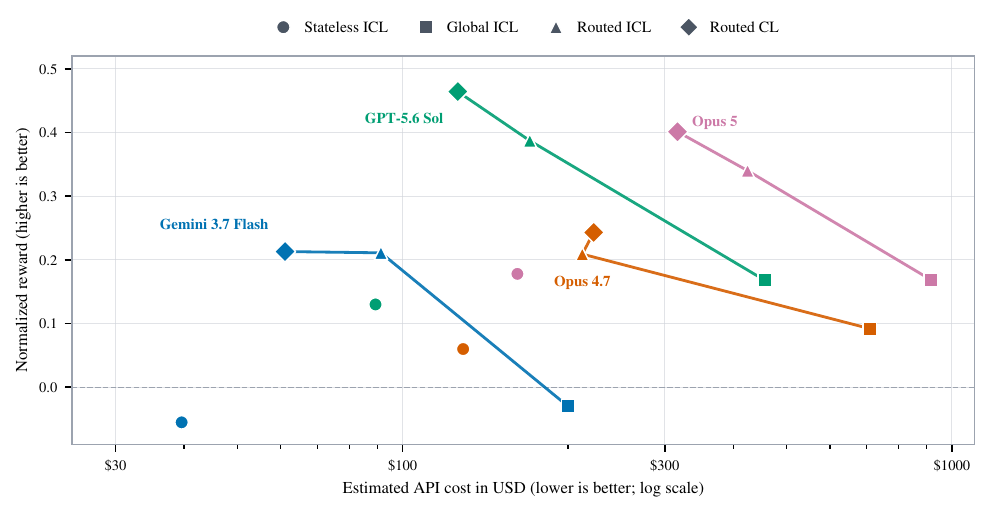}
  \caption{Mean normalized reward versus mean estimated API cost across the five
  matched rollouts in Table~\ref{tab:main-results}. Estimated costs include every
  model call under the frozen pricing schedule. Within each model, lines connect
  Global ICL, Routed ICL, and Routed CL; Stateless ICL appears separately.
  Table~\ref{tab:api-costs} reports the exact mean costs. \textbf{Routed CL
  achieves the highest mean reward among the four staged conditions for every
  model and lies on the empirical reward--cost Pareto frontier.}}
  \label{fig:reward-cost-tradeoff}
\end{figure}

\runin{Routed CL improves the reward--cost trade-off.}
Among the four staged
conditions, Routed CL lies on the empirical reward--cost Pareto frontier for
every model (Figure~\ref{fig:reward-cost-tradeoff}). It achieves higher reward than Global ICL at 65--72\% lower cost
and costs 25--33\% less than Routed ICL on three models; on Opus 4.7, it costs
5\% more for a 0.034 reward gain. Routed CL costs more than ReasoningBank:
relative to Global RB, it gains 0.189--0.358 reward points at 31--44\% higher
cost and achieves higher reward than Routed RB for every model
(Figure~\ref{fig:reasoning-bank-tradeoff}).

\begin{figure}[H]
  \centering
  \includegraphics[width=\linewidth]{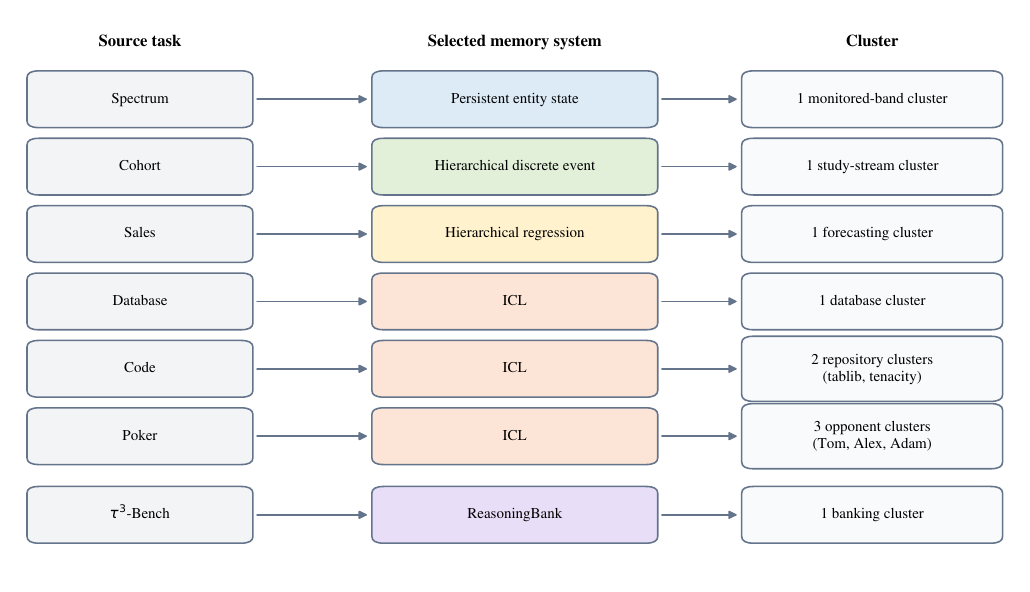}
  \caption{Ground-truth assignment used for diagnostic
  comparisons. The mapping was fixed before confirmatory evaluation and is not
  an optimized policy or upper bound. It shows the joint memory-system and
  cluster assignment for each source task.}
  \label{fig:ground-truth-assignment}
\end{figure}

\runin{Routing leaves model-dependent headroom.}
Replacing the LLM router with Ground-truth assignment, a fixed source-informed
diagnostic, raises normalized reward by only 0.008 for Opus 5 and 0.014 for
GPT-5.6 Sol, but by 0.069 for Opus 4.7 and 0.071 for Gemini 3.7 Flash. Across 20
model--rollout combinations, the router independently chooses the same memory
system in 86\% of source-task cases, the same cluster partition in 91\%, and both
in 79\% (Table~\ref{tab:routing-assignments}).
Figure~\ref{fig:ground-truth-assignment} visualizes the coupled Ground-truth
assignment. Disagreements concentrate on Code for Gemini and Opus 4.7, which
often select ReasoningBank rather than ICL, and on Sales for Gemini, which never
selects hierarchical regression. Routed CL also trails Routed ICL on these
pairs.


\section{Conclusion and Future Work}

Agents learning across heterogeneous tasks must decide how to represent and
group relevant experiences. We introduce Routed Continual Learning, which makes
these decisions jointly by selecting a memory system and a stateful cluster for
each agent run.

On our heterogeneous stream spanning CL-Bench and $\tau^3$-Bench, Routed CL
achieves higher reward than both Global ICL and Routed ICL for all four
language models at substantial cost reductions (e.g., saving 65--72\% compared to Global ICL). Notably, Global ICL, found to be the strongest performing system by the original CL-Bench when evaluated on isolated tasks, gains almost nothing (mean +0.015 in aggregated normalized reward) relative to the stateless condition when the tasks are interleaved. These results support treating continual learning as a routing problem that
jointly determines how experience is represented and which instances share
state.


Our experiments process instances sequentially, and each rollout uses one
language model as the actor and for all model-mediated memory-system roles.
Although the architecture isolates cluster state and defines atomic reads and
writes, we do not evaluate simultaneous updates, stale reads, or cross-model
reuse of cluster state. These remain important tests for distributed deployment.

Future work could study richer interactions between components: reading from
multiple clusters for a single instance, and updating several clusters or
memory-system representations from the resulting trajectory, while controlling
cost and interference.

\section*{AI Use Statement}

We used generative AI tools to help refine hypotheses and conceptual framing, provide feedback on experimental design, implement and debug the proposed methods and experimental infrastructure, inspect and compile experimental results, search relevant literature, prepare figures, and draft and revise portions of the manuscript. Large language models also serve as experimental actors, routers, and information extractors in the evaluated systems; those uses are described separately in the System Components and Experiments sections. The authors made all final research and writing decisions, reviewed and revised all AI-assisted text, verified citations against primary sources, tested AI-assisted code, and checked reported results against the underlying experimental records.

\section*{Ethics Statement}

Our experiments use previously released benchmarks and simulated interactions. The
motivating example in the Introduction highlights a capability risk:
pooling experience can enable a group of agents to achieve outcomes beyond the
capabilities of any individual agent, including in security-sensitive settings.
Persistent state can also propagate sensitive, erroneous, or adversarial
information. Deployments should therefore enforce task- and user-level
authorization, provenance, retention and deletion policies, and isolation
across security boundaries, while assessing the capabilities of the combined
system rather than only its component models.

\section*{Reproducibility Statement}

The System Components and Experiments sections specify the architecture, memory-system library, routing and adapter procedures, baselines, metrics, and matched-rollout protocol. The appendix reproduces the fixed prompts and memory-system descriptions and documents adapter validation, benchmark adaptations, environment isolation, the frozen evaluation contract, and cost accounting.

\bibliographystyle{assets/plainnat}
\bibliography{bibliography}

\clearpage
\newpage
\beginappendix

\section{Additional Experimental Details}\label{app:experiments}

\subsection{Source Tasks}

\begin{table}[H]
  \centering
  \footnotesize
  \setlength{\tabcolsep}{4pt}
  \renewcommand{\arraystretch}{1.08}
  \caption{Source tasks used to construct the heterogeneous evaluation stream.}
  \label{tab:clbench-tasks}
  \begin{tabularx}{\linewidth}{@{}>{\raggedright\arraybackslash}p{0.19\linewidth}r>{\raggedright\arraybackslash}p{0.30\linewidth}X@{}}
    \toprule
    Task & Instances & Instance objective & Reusable cross-instance information \\
    \midrule
    Codebase Adaptation & 19 & Resolve repository issues under a command budget & Repository structure, test workflows, and contributor conventions \\
    Cohort Studies & 20 & Estimate event-free curves for 36 population cohorts & Population structure across partially observed and differently sampled studies \\
    Database Exploration & 40 & Answer database questions using limited SQL queries & Schema, joins, encodings, and migration-induced changes \\
    Exploitable Poker & 120 & Maximize profit over independent poker hands & Opponent behavior and changes in strategy \\
    Sales Prediction & 12 & Forecast five years of product--location demand & Shared growth, product and location effects, and forecast calibration \\
    Blind Spectrum Monitoring & 90 & Infer long-run transmitter occupancy from noisy scans & Persistent transmitters observed intermittently across scans \\
    $\tau^3$-Bench Banking & 97 & Resolve policy-grounded banking requests through multi-turn conversation and tool use & Procedures, policy interpretation, and failure-avoidance strategies \\
    \midrule
    \textbf{Total} & \textbf{398} & & \\
    \bottomrule
  \end{tabularx}
\end{table}

\subsection{Heterogeneous Stream Construction}
\label{app:stream-construction}

Feedback produced by a source task that advances automatically is buffered and
delivered when that source task is next activated. This preserves its causal
position without attaching it to another task's trajectory.

\subsection{Implementation and Rollout Configuration}
\label{app:rollout-provenance}

The headline table aggregates five matched stream seeds (42--46) per model.
Ground-truth assignment also reports five matched rollouts for every model.

\subsection{Benchmark Protocol Modifications}

\subsubsection{Sales Prediction Workspace Isolation}
\label{app:sales-isolation}

The released Sales Prediction runtime preserves its writable execution
environment across instances and explicitly encourages the actor to save code,
models, and notes under \texttt{/app}. This provides a general-purpose
cross-instance storage channel outside the continual-learning system being
evaluated.

For every method in our evaluation, we instead discard the Sales container at
each instance boundary and start a fresh one. The harness then reconstructs the
current instance's data room. Benchmark-provided observations, including the
currently visible historical data and previous-round feedback, remain available
through the native task interface; actor-created files do not persist. The task
instructions are changed accordingly so that they accurately describe the
execution environment. This isolation strengthens attribution to the evaluated
continual-learning system but makes our Sales results not strictly
protocol-equivalent to historical CL-Bench results.

\subsubsection{Codebase Adaptation Isolation}
\label{app:code-isolation}

CL-Bench describes Codebase Adaptation as beginning from an issue's base commit
and evaluating the submitted patch against hidden regression tests. The CL-Bench
1.0 release commit (\texttt{a7ddacb}) did not fully enforce this separation.
Actor workspaces retained the shared image's Git database, from which later fix
commits could be inspected, and the official regression-test patch was applied
before the actor began interacting with the repository.

Our hardened actor initialization first restores the base commit and removes
untracked and ignored build artifacts. It then removes the inherited Git database
and creates a fresh repository containing only a single snapshot commit. The
official regression-test patch is never applied in the actor container. Evaluation
remains unchanged: after submission, a separate evaluator reconstructs the base
tree, applies the hidden regression-test patch and the actor's submitted patch,
and runs the designated tests. The runtime records whether actor sanitization was
enabled and whether the test patch was visible for every run.

This change matches the benchmark's stated hidden-test protocol but differs from
the released runtime used by historical runs. Aggregate results alone do not show
whether a historical agent used either leakage channel. We therefore report our
Code results as sanitized evaluations and avoid treating comparisons with
historical Code scores as strictly equivalent.

\subsection{Frozen Evaluation Contract}
\label{app:evaluation-contract}

Before confirmatory runs, we freeze the router prompt and memory-system
descriptions; all memory-system implementations; the prompts and schemas used for
adapter configuration, typed extraction, ReasoningBank self-judgment, and memory
extraction; statistical defaults; the Ground-truth assignment mapping; and the
task harness and scoring code. We do not revise these components in response to
confirmatory outcomes. Within each
matched comparison, systems receive the same native instance prompts, tools,
response schemas, visible feedback, action budgets, and actor context-window and
response limits.

When deciding how much retained ICL history fits, the runtime reserves 500 tokens
within each configured 100,000-token budget. If the remaining history exceeds
that allowance, it removes the oldest retained messages first. Global ICL applies
this policy to one stream-wide history; Routed ICL and ICL clusters selected by
Routed CL apply it independently per cluster.

Within each model--condition pair, the same evaluated model serves as the actor
and, where applicable, powers selection, adapter configuration, typed extraction,
ReasoningBank self-judgment, and memory extraction. Resource measurements include
all of these calls.

\subsection{Metric Definitions}
\label{app:metric-definitions}

For a fixed language model and rollout seed, let $\bar r_d(M)$ denote the mean
native reward of method $M$ on source task $d$.

\paragraph{Normalized reward.}
For the six CL-Bench tasks, we follow CL-Bench and normalize each task's reward
using its external reference $r_d^{\mathrm{ext}}$, the reported stateless
GPT-5.4 ICL reward for task $d$~\citep{asawa2026clbench}:
\[
\frac{\bar r_d(M)-r_d^{\mathrm{ext}}}
     {r_d^{\max}-r_d^{\mathrm{ext}}}.
\]
Here, $r_d^{\max}$ is the benchmark-defined maximum reward for task $d$. For
$\tau^3$-Bench, whose native reward already lies in $[0,1]$, we use that reward
directly.

\paragraph{Stateless-reference gain.}
Stateless-reference gain instead uses Stateless ICL $B$ with the same language model
and rollout seed:
\[
\frac{\bar r_d(M)-\bar r_d(B)}
     {r_d^{\max}-\bar r_d(B)}.
\]
Both metrics are averaged equally across the seven source tasks. These task-level
normalizations place heterogeneous rewards on a common scale; they do not make
our mixed-stream and hardened protocols strictly comparable with historical
aggregate CL-Bench results.

\subsection{API Cost Accounting}
\label{app:cost-accounting}

We include every model call used for actor decisions, selection, adapter
configuration, typed extraction, ReasoningBank self-judgment, and memory
extraction. Costs use a frozen, dated public list-price schedule. Uncached input,
cache-read, cache-write, and output tokens are priced separately, and reasoning
tokens reported separately by a provider are priced as output. This policy
captures the caching advantage of long ICL histories.

\section{Additional Method Details}
\label{app:method-details}

\subsection{Adapter Validation and Evidence Repair}
\label{app:adapter-validation}

ICL uses a fixed interface, whereas ReasoningBank and the numerical memory
systems use cluster-specific adapters. ReasoningBank's adapter identifies the
boundary between a repeated prompt wrapper and its changing request; its memory
induction and retrieval behavior remain fixed. Numerical adapters instead bind
visible task fields to roles such as targets, identities, groups, and time, and
choose among the configuration options permitted by each memory system's fixed
typed schema.

A configurable cluster's first instance receives the original prompt unchanged.
Its completed trajectory becomes the warm-up example, and adapter configuration
occurs only if the cluster is subsequently reused.

On the first reuse of a configurable numerical cluster, Context Memory Adapter
configures its adapter from the completed warm-up trajectory together with the
current visible prompt and response schema. A fresh memory-system runtime replays the
warm-up trajectory, prepares the current instance, and checks the resulting state
and bindings. If validation fails, Context Memory Adapter returns the deterministic
error to the configuring LLM, which may return one complete repair. A second
failure terminates adapter configuration rather than accepting an unvalidated
configuration. Successful validation initializes the memory-system state from the
warm-up trajectory, and the resulting adapter configuration is frozen for the
lifetime of the cluster.

After configuration, numerical memory systems invoke a typed extractor whenever new
evidence becomes available. The extractor receives one complete visible event,
such as the current query or a completed trajectory, and emits the records
required by the configured adapter. Memory-system-specific validators check their
schema, internal consistency, and exact provenance against the source event. An
invalid batch receives at most one validation-informed repair; persistent failure
is reported rather than accepted or silently truncated. The extractor does not
fit the statistical model, associate latent entities, or generate predictions.
Those operations remain inside the fixed memory-system runtime.

\subsection{Numerical Memory-System Implementations}
\label{app:numerical-operators}

\paragraph{Hierarchical Regression.}
This memory system models repeated numerical outcomes indexed by stable keys,
categorical factors, and optionally time. Its adapter chooses the target family
and temporal structure, while the extractor establishes consistent dimension
names and emits requested prediction points and visible realized outcomes. The
runtime pools accumulated evidence through hierarchical regression implemented
in PyMC, with global, key-specific, factor, and temporal terms, refitting as new
outcomes arrive. Once an estimate has repeated support, it renders a compact
forecast with its fitted observation-noise estimate and recent observations;
the actor may still revise that estimate using the current environment.

\paragraph{Hierarchical Discrete-Event State.}
This memory system models monotone event-free probabilities over ordered intervals
through conditional hazards. Its adapter chooses the number of intervals and
stable categorical context roles; the extractor maps visible aggregate
measurements to target identities, reusable features, context levels,
probability curves, and support weights. A hierarchical discrete-hazard model
implemented in PyMC pools an interval-specific baseline with feature-, entity-,
and context-level effects, then converts predicted hazards back into valid
monotone curves. The runtime ingests visible aggregates from completed
trajectories and renders the resulting curves for requested targets as advisory
context on subsequent instances.

\paragraph{Persistent-Entity State.}
This memory system handles variable-size, unordered measurement sets whose underlying
entities may persist without stable identifiers. Its adapter identifies the
numerical coordinates that define entity identity, visible measurements that
determine their uncertainty scales, values to predict, target family, and any
optional query window. The extractor copies each visible measurement set into
typed records without performing association. The runtime links observations
through uncertainty-aware matching, updates persistent identity estimates and
online Bayesian regressions, and renders current or previously confirmed
entities relevant to the query as advisory context.

\section{Additional Results}
\label{app:additional-results}

\subsection{Stateless-Reference Gain}

\begin{table}[H]
  \centering
  \footnotesize
  \setlength{\tabcolsep}{3pt}
  \renewcommand{\arraystretch}{1.16}
  \caption{Gain relative to matched model-specific Stateless ICL across five
  rollouts. Each cell reports the mean, with $[\min,\max]$ beneath. Higher is
  better; bold marks the best of the four staged conditions.}
  \label{tab:fixed-reference-gain}
  \begin{tabularx}{\linewidth}{@{}l*{3}{>{\centering\arraybackslash}X}@{\hspace{5pt}\vrule width 0.35pt\hspace{5pt}}>{\centering\arraybackslash}X@{}}
    \toprule
    Model & Global ICL & Routed ICL & Routed CL & \shortstack{Ground-truth\\assignment} \\
    \midrule
    Gemini 3.7 Flash & \shortstack{0.019\\{\scriptsize$[0.001,0.038]$}} & \shortstack{0.229\\{\scriptsize$[0.193,0.271]$}} & \shortstack{\textbf{0.265}\\{\scriptsize$[0.150,0.341]$}} & \shortstack{0.313\\{\scriptsize$[0.281,0.356]$}} \\
    GPT-5.6 Sol & \shortstack{0.043\\{\scriptsize$[-0.010,0.074]$}} & \shortstack{0.307\\{\scriptsize$[0.252,0.357]$}} & \shortstack{\textbf{0.381}\\{\scriptsize$[0.364,0.431]$}} & \shortstack{0.401\\{\scriptsize$[0.352,0.453]$}} \\
    Opus 4.7 & \shortstack{0.028\\{\scriptsize$[0.017,0.037]$}} & \shortstack{0.159\\{\scriptsize$[0.137,0.201]$}} & \shortstack{\textbf{0.207}\\{\scriptsize$[0.147,0.281]$}} & \shortstack{0.273\\{\scriptsize$[0.236,0.332]$}} \\
    Opus 5 & \shortstack{$-0.029$\\{\scriptsize$[-0.084,0.007]$}} & \shortstack{0.193\\{\scriptsize$[0.151,0.279]$}} & \shortstack{\textbf{0.265}\\{\scriptsize$[0.237,0.318]$}} & \shortstack{0.270\\{\scriptsize$[0.256,0.283]$}} \\
    \bottomrule
  \end{tabularx}
\end{table}

\begin{table}[H]
  \centering
  \footnotesize
  \setlength{\tabcolsep}{6pt}
  \renewcommand{\arraystretch}{1.12}
  \caption{Paired normalized-reward differences between Routed CL and Routed
  ICL across the five matched stream seeds. The second column reports the mean
  with $[\min,\max]$ beneath; positive values favor Routed CL.}
  \label{tab:paired-routing-differences}
  \vspace{6pt}
  \begin{tabular}{@{}lcc@{}}
    \toprule
    Model & \shortstack{Routed CL $-$ Routed ICL\\Mean $[\min,\max]$} & \shortstack{Positive\\rollouts} \\
    \midrule
    Gemini 3.7 Flash & \shortstack{0.002\\{\scriptsize$[-0.090,0.073]$}} & 4/5 \\
    GPT-5.6 Sol & \shortstack{0.076\\{\scriptsize$[0.060,0.110]$}} & 5/5 \\
    Opus 4.7 & \shortstack{0.034\\{\scriptsize$[-0.009,0.085]$}} & 4/5 \\
    Opus 5 & \shortstack{0.061\\{\scriptsize$[0.044,0.072]$}} & 5/5 \\
    \bottomrule
  \end{tabular}
\end{table}

\subsection{Comparison with ReasoningBank}
\label{app:reasoning-bank-control}

\begin{figure}[H]
  \centering
  \includegraphics[width=\linewidth]{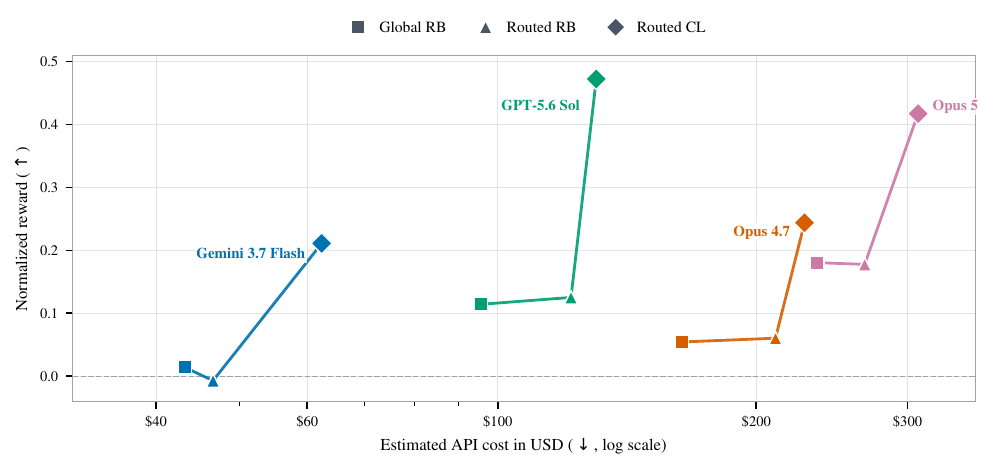}
  \caption{Mean normalized reward versus mean estimated API cost for the
  ReasoningBank comparison over three matched rollouts. Within each model, lines
  connect Global RB, Routed RB, and Routed CL. Routed CL is recomputed over these
  three rollouts and therefore differs from the five-rollout values in
  Table~\ref{tab:main-results}. Points toward the upper left are better.}
  \label{fig:reasoning-bank-tradeoff}
\end{figure}

To compare against a prior memory system and separate cluster routing from
memory-system choice, we evaluate our documented ReasoningBank
adaptation~\citep{ouyang2026reasoningbank} under two fixed-system conditions.
\textbf{Global RB} maintains one ReasoningBank state across the complete stream,
whereas \textbf{Routed RB} uses the LLM router to partition instances among
ReasoningBank clusters. Section~\ref{sec:method} describes how this adaptation
preserves ReasoningBank's core retrieval, extraction, and consolidation loop and
where it differs from the original implementation.
Figure~\ref{fig:reasoning-bank-tradeoff} compares these conditions and Routed CL
over the same three matched rollouts. The path from Global RB to Routed RB shows
the effect of routing a fixed alternative memory system; the final segment to
Routed CL shows the benefit of access to the full memory-system library under
routed execution.

\subsection{Learning Dynamics}

\begin{figure}[H]
  \centering
  \includegraphics[width=\linewidth]{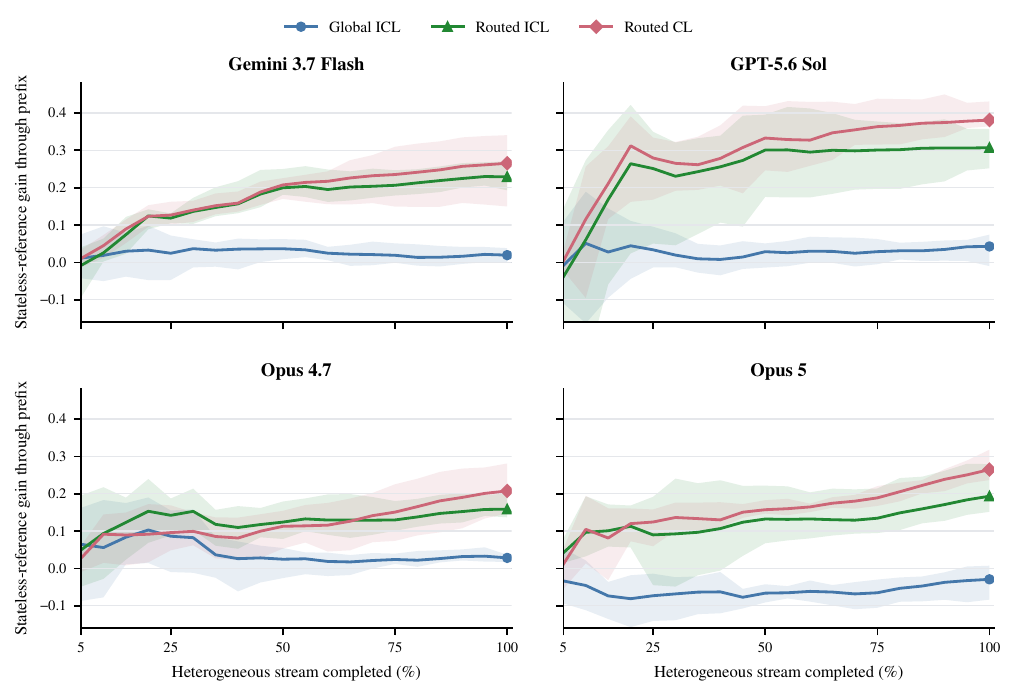}
  \caption{Stateless-reference gain through successive prefixes of the heterogeneous
  stream. Lines show means across five matched rollouts and shading spans
  $[\min,\max]$; the endpoints reproduce the full-stream gains in
  Table~\ref{tab:fixed-reference-gain}.}
  \label{fig:learning-curves}
\end{figure}

Figure~\ref{fig:learning-curves} examines whether the benefit of persistent
state grows as experience accumulates. At each stream prefix, we compute each
method's reward difference from its matched Stateless ICL rollout separately
for every source task, normalize by that task's full-rollout headroom, and
average the seven tasks equally. Curves begin at 5\%. Before a source task first
appears within a prefix, its contribution is defined as zero; this preserves a
fixed seven-task average rather than averaging over a changing set of tasks.
Early points are closer to zero because source tasks that have not yet appeared
contribute zero. This affects every method equally within a matched rollout, and
all seven source tasks have appeared by 15\% of the stream.

\subsection{Estimated API Cost}

\begin{table}[H]
  \centering
  \footnotesize
  \setlength{\tabcolsep}{3pt}
  \renewcommand{\arraystretch}{1.16}
  \caption{Mean estimated API cost in USD per 398-instance rollout across five
  matched seeds. Costs include every model call under the accounting policy in
  Appendix~\ref{app:cost-accounting}; Table~\ref{tab:token-usage} reports the
  corresponding token breakdown. Lower is better.}
  \label{tab:api-costs}
  \begin{tabularx}{\linewidth}{@{}l*{4}{>{\centering\arraybackslash}X}@{\hspace{5pt}\vrule width 0.35pt\hspace{5pt}}>{\centering\arraybackslash}X@{}}
    \toprule
    Model & Stateless ICL & Global ICL & Routed ICL & Routed CL & \shortstack{Ground-truth\\assignment} \\
    \midrule
    Gemini 3.7 Flash & \$39.59 & \$200.14 & \$91.31 & \$61.08 & \$59.20 \\
    GPT-5.6 Sol & \$89.24 & \$456.22 & \$170.40 & \$125.99 & \$117.08 \\
    Opus 4.7 & \$128.86 & \$708.78 & \$212.37 & \$222.72 & \$212.04 \\
    Opus 5 & \$161.77 & \$914.36 & \$424.54 & \$316.61 & \$265.10 \\
    \bottomrule
  \end{tabularx}
\end{table}

\clearpage
\subsection{Token Usage}

\begin{table}[H]
  \centering
  \scriptsize
  \setlength{\tabcolsep}{3.2pt}
  \renewcommand{\arraystretch}{1.05}
  \caption{Mean token usage (millions) per 398-instance rollout across the same
  five matched rollouts as Table~\ref{tab:api-costs}. The three input columns
  partition total input; cached input is not counted again in the total. Lower
  is better, and comparisons are most direct within a model because tokenizers
  differ.}
  \label{tab:token-usage}
  \begin{tabularx}{\linewidth}{@{}ll*{5}{>{\centering\arraybackslash}X}@{}}
    \toprule
    Model & Condition & \shortstack{Fresh\\input} & \shortstack{Cache-read\\input} & \shortstack{Cache-write\\input} & Output & \shortstack{Usage\\total} \\
    \midrule
    Gemini 3.7 Flash & Stateless ICL & 34.1 & 91.1 & 0.0 & 0.7 & 127.2 \\
     & Global ICL & 237.4 & 192.8 & 0.0 & 0.7 & 432.2 \\
     & Routed ICL & 101.1 & 137.3 & 0.0 & 0.6 & 239.7 \\
     & Routed CL & 58.8 & 105.5 & 0.0 & 0.9 & 166.7 \\
     & \shortstack[l]{GT assignment} & 57.7 & 101.3 & 0.0 & 0.9 & 161.2 \\
    \cmidrule(lr){2-7}
    GPT-5.6 Sol & Stateless ICL & 10.2 & 79.0 & 0.0 & 0.8 & 90.1 \\
     & Global ICL & 90.6 & 191.2 & 0.0 & 0.9 & 282.6 \\
     & Routed ICL & 25.9 & 133.5 & 0.0 & 0.7 & 160.1 \\
     & Routed CL & 16.8 & 94.9 & 0.0 & 1.0 & 112.8 \\
     & \shortstack[l]{GT assignment} & 14.1 & 100.0 & 0.0 & 1.0 & 115.2 \\
    \cmidrule(lr){2-7}
    Opus 4.7 & Stateless ICL & 6.2 & 79.1 & 5.9 & 0.9 & 92.0 \\
     & Global ICL & 2.6 & 169.3 & 93.7 & 1.0 & 266.7 \\
     & Routed ICL & 8.0 & 98.7 & 16.1 & 0.9 & 123.7 \\
     & Routed CL & 15.8 & 91.5 & 11.0 & 1.2 & 119.4 \\
     & \shortstack[l]{GT assignment} & 14.8 & 98.5 & 9.6 & 1.2 & 124.1 \\
    \cmidrule(lr){2-7}
    Opus 5 & Stateless ICL & 5.4 & 109.7 & 7.4 & 1.4 & 123.8 \\
     & Global ICL & 3.3 & 231.1 & 117.0 & 2.0 & 353.5 \\
     & Routed ICL & 3.4 & 199.7 & 43.7 & 1.4 & 248.1 \\
     & Routed CL & 12.6 & 156.1 & 21.3 & 1.7 & 191.7 \\
     & \shortstack[l]{GT assignment} & 16.6 & 125.2 & 12.4 & 1.7 & 156.0 \\
    \bottomrule
  \end{tabularx}
  \vspace{2pt}
  \begin{minipage}{\linewidth}
    \scriptsize
    \emph{Accounting note.} Fresh input excludes cache reads and writes. Usage
    total includes all reported input, cache, output, and reasoning tokens without
    double-counting; providers differ in whether reasoning tokens appear in the
    Output column.
  \end{minipage}
\end{table}
\FloatBarrier

\subsection{Performance by Source Task}

Gemini's 2.41$\times$ Poker cost ratio occurs in the three rollouts where Routed
CL merged all opponents into one ICL cluster
(Table~\ref{tab:routing-assignments}, panel~b). The merged cluster required the actor to
repeatedly process a longer history; costs were comparable to Routed ICL in the
two opponent-partitioned rollouts.

\subsection{Memory-System and Cluster Assignments}

Table~\ref{tab:routing-assignments} reports the router's memory-system assignments
in its upper panel and the corresponding cluster partitions in its lower panel.

\begin{table}[H]
  \centering
  \footnotesize
  \setlength{\tabcolsep}{4pt}
  \renewcommand{\arraystretch}{1.08}
  \caption{Memory-system and cluster assignments across four models and five
  matched rollouts per model. Panel (a) counts assignments to each memory system;
  panel (b) lists the number of clusters in rollout order. Bold entries match the
  corresponding component of Ground-truth assignment; cluster agreement requires
  the complete partition, not only its size. Source-task identities and grouping
  keys are used only for post-hoc evaluation. Memory-system abbreviations follow
  Table~\ref{tab:operator-library}.}
  \label{tab:routing-assignments}
  \begin{tabularx}{\linewidth}{@{}l>{\centering\arraybackslash}p{0.13\linewidth}*{4}{>{\centering\arraybackslash}X}@{}}
    \toprule
    Source task & Ground truth & \shortstack{Gemini 3.7\\Flash} & GPT-5.6 Sol & Opus 4.7 & Opus 5 \\
    \midrule
    \multicolumn{6}{@{}l}{\textbf{(a) Memory-system assignment}} \\
    \addlinespace[2pt]
    Code & ICL & RB 4, \textbf{ICL 1} & \textbf{ICL 5} & RB 3, \textbf{ICL 2} & \textbf{ICL 4}, RB 1 \\
    Cohort & HDE & \textbf{HDE 5} & \textbf{HDE 5} & \textbf{HDE 4}, RB 1 & \textbf{HDE 5} \\
    Database & ICL & \textbf{ICL 5} & \textbf{ICL 5} & \textbf{ICL 3}, RB 2 & \textbf{ICL 5} \\
    Poker & ICL & \textbf{ICL 5} & \textbf{ICL 5} & \textbf{ICL 5} & \textbf{ICL 5} \\
    Sales & HR & ICL 4, RB 1 & \textbf{HR 5} & \textbf{HR 4}, ICL 1 & \textbf{HR 5} \\
    Spectrum & PES & \textbf{PES 5} & \textbf{PES 5} & \textbf{PES 5} & \textbf{PES 5} \\
    $\tau^3$-Bench & RB & \textbf{RB 5} & \textbf{RB 5} & \textbf{RB 4}, ICL 1 & \textbf{RB 3}, ICL 2 \\
    \midrule
    \textbf{Memory-system agreement} & & \textbf{26/35} & \textbf{35/35} & \textbf{27/35} & \textbf{32/35} \\
    \midrule
    \multicolumn{6}{@{}l}{\textbf{(b) Cluster assignment}} \\
    \addlinespace[2pt]
    Code & Repository & $1,1,\mathbf{2},1,\mathbf{2}$ & $\mathbf{2},\mathbf{2},\mathbf{2},\mathbf{2},\mathbf{2}$ & $\mathbf{2},\mathbf{2},\mathbf{2},\mathbf{2},\mathbf{2}$ & $\mathbf{2},\mathbf{2},\mathbf{2},\mathbf{2},\mathbf{2}$ \\
    Cohort & Stream & $\mathbf{1},\mathbf{1},\mathbf{1},\mathbf{1},\mathbf{1}$ & $\mathbf{1},\mathbf{1},\mathbf{1},\mathbf{1},\mathbf{1}$ & $\mathbf{1},\mathbf{1},\mathbf{1},\mathbf{1},\mathbf{1}$ & $\mathbf{1},\mathbf{1},\mathbf{1},\mathbf{1},\mathbf{1}$ \\
    Database & Stream & $\mathbf{1},\mathbf{1},\mathbf{1},\mathbf{1},\mathbf{1}$ & $\mathbf{1},\mathbf{1},\mathbf{1},\mathbf{1},\mathbf{1}$ & $\mathbf{1},\mathbf{1},\mathbf{1},\mathbf{1},\mathbf{1}$ & $\mathbf{1},\mathbf{1},\mathbf{1},\mathbf{1},\mathbf{1}$ \\
    Poker & Opponent & $1,1,\mathbf{3},1,\mathbf{3}$ & $\mathbf{3},\mathbf{3},\mathbf{3},\mathbf{3},\mathbf{3}$ & $\mathbf{3},2,\mathbf{3},\mathbf{3},\mathbf{3}$ & $\mathbf{3},\mathbf{3},\mathbf{3},\mathbf{3},\mathbf{3}$ \\
    Sales & Stream & $\mathbf{1},\mathbf{1},\mathbf{1},\mathbf{1},\mathbf{1}$ & $\mathbf{1},\mathbf{1},\mathbf{1},\mathbf{1},\mathbf{1}$ & $\mathbf{1},\mathbf{1},\mathbf{1},\mathbf{1},\mathbf{1}$ & $\mathbf{1},\mathbf{1},\mathbf{1},\mathbf{1},\mathbf{1}$ \\
    Spectrum & Stream & $\mathbf{1},\mathbf{1},\mathbf{1},\mathbf{1},\mathbf{1}$ & $\mathbf{1},\mathbf{1},\mathbf{1},\mathbf{1},\mathbf{1}$ & $\mathbf{1},\mathbf{1},\mathbf{1},\mathbf{1},\mathbf{1}$ & $\mathbf{1},\mathbf{1},\mathbf{1},\mathbf{1},\mathbf{1}$ \\
    $\tau^3$-Bench & Stream & $\mathbf{1},\mathbf{1},\mathbf{1},\mathbf{1},\mathbf{1}$ & \shortstack{$\mathbf{1},27,25,$\\$23,25$} & \shortstack{$\mathbf{1},23,86,$\\$\mathbf{1},\mathbf{1}$} & $\mathbf{1},\mathbf{1},\mathbf{1},\mathbf{1},\mathbf{1}$ \\
    \midrule
    \textbf{Cluster agreement} & & \textbf{29/35} & \textbf{31/35} & \textbf{32/35} & \textbf{35/35} \\
    \midrule
    \textbf{Joint agreement} & & \textbf{23/35} & \textbf{31/35} & \textbf{25/35} & \textbf{32/35} \\
    \bottomrule
  \end{tabularx}
\end{table}

``Repository'' denotes separate \texttt{tablib} and \texttt{tenacity} clusters,
``Opponent'' one cluster for each of the three Poker opponents, and ``Stream'' one
cluster containing all instances from that source task. Gemini 3.7 Flash merges
the two Code repositories in three of five rollouts and the three Poker opponents
in three of five. GPT-5.6 Sol fragments $\tau^3$-Bench into 23--27 clusters in
four of five rollouts, while Opus 4.7 creates 23 and 86 clusters in two rollouts.
These are descriptive routing errors; the table does not attribute reward
differences to either failure mode.

\subsection{Ground-truth Assignment}

\section{Prompts and Memory-System Descriptions}
\label{app:prompts}

This appendix reproduces the fixed natural-language instructions that define
Routed CL's experimental method: the router prompt and memory-system
descriptions, the prompts Context Memory Adapter uses for adapter configuration
and typed extraction, ReasoningBank self-judgment and memory extraction, and the
fixed wrappers used to expose learned state to the actor.
Line wrapping and runtime-value placeholders are normalized, but the fixed
wording is otherwise verbatim. Text in angle brackets denotes visible data
inserted at runtime. We omit native
benchmark instance prompts and response schemas, which are benchmark inputs
rather than method prompts, as well as provider-specific transport and retry
messages that only enforce an already specified structured-output schema.

Stateless ICL and Global ICL add no system instruction to the native benchmark
prompt. Routed ICL uses the router below, but its actor likewise receives only
the native prompt and the visible history assigned to its cluster.

\subsection{Router}
\label{app:selector-prompt}

The router receives the following system prompt before every instance. In the
verbatim implementation prompt, \emph{task} denotes the current instance and
\emph{system} denotes a candidate memory system.

\begin{lstlisting}[style=prompt]
You route tasks among available continual-learning systems.
Each existing cluster owns one fixed system and state accumulated from earlier
tasks.

Use only the supplied task, cluster summaries, and system declarations. Infer
what information could usefully transfer to future related tasks. When creating
a cluster, select the system whose declared learning behavior and requirements
best match that information. Do not assume undeclared capabilities or prefer a
system because of its identifier or position in the candidate list.

Reuse an existing cluster only when its description indicates that its state
applies to the same scope and is likely to help without conflict. Otherwise
create a new cluster. Multiple clusters may use the same system. Describe the
scope of future tasks that should benefit from a new cluster.

Every structured response must include all four fields. Use null for fields that
do not apply:
- Existing: {"kind":"existing","cluster_id":"cluster-0001","system_id":null,
  "description":null}
- New: {"kind":"new","cluster_id":null,"system_id":"system-id",
  "description":"Concise scope for future tasks"}

Return only the structured selection.
\end{lstlisting}

The user message is a JSON object with the following structure. The current
instance prompt and response schema are copied from the native interface; no
source-task identity is included.

\begin{lstlisting}[style=prompt]
{
  "clusters": [
    {
      "description": <immutable cluster description>,
      "id": <cluster identifier>,
      "system_id": <system identifier>,
      "trajectory_count": <number of completed trajectories>
    }, ...
  ],
  "systems": [
    {"description": <system description>, "id": <system identifier>}, ...
  ],
  "task": {
    "prompt": <current visible instance prompt>,
    "response_schema": <current structured-response schema>
  }
}
\end{lstlisting}

The frozen system descriptions supplied in the Routed CL condition are:

\begin{lstlisting}[style=prompt]
icl:
What: Bounded verbatim episodic memory that retains prior actor-visible
interactions in temporal order within its configured context window. Choose
when: Future related tasks continue the same entities, environment, or evolving
state, or when exact prior facts, actions, observations, or outcomes may be
needed. Avoid when only compact task-independent strategies need to transfer.

reasoning_bank_v1:
What: Self-judged reasoning memory that distills successful and failed
actor-visible experiences into compact strategies and preventative lessons.
For each new task, it retrieves the memories attached to the single most
semantically similar prior experience.
Choose when: Tasks are self-contained episodes that benefit primarily from
reusable procedures, decision patterns, or failure-avoidance lessons distilled
from related prior experiences.

hierarchical_regression_v1:
What: Hierarchical online regression over keyed scalar outcomes. Choose when:
The reusable target is a scalar numeric outcome observed repeatedly for keyed
entities or panels over time. Scalar numeric outcomes with keyed dimensions,
observation times, requested prediction times, and provenance.

hierarchical_discrete_event_v1:
What: A hierarchical discrete-event model for monotone event-free probability
curves. Choose when: The reusable target is a jointly modeled monotone
probability vector over fixed ordered intervals, pooled across groups, features,
or entities. Monotone probability curves with categorical groups, features,
entity keys, weights, and provenance.

persistent_entity_state_v1:
What: Online regression over unordered measurements of persistent latent
entities. Choose when: Each step contains a variable-size set of noisy detections
that may correspond to persistent entities. Raw numeric identity coordinates,
natural scale references, numeric values, optional visible bounds, step index,
and provenance.
\end{lstlisting}

The router returns a structured object with four fields:
\texttt{kind}, \texttt{cluster\_id}, \texttt{system\_id}, and
\texttt{description}.

\subsection{Adapter Configuration}
\label{app:adapter-configuration}

ICL requires no cluster-specific adapter configuration. ReasoningBank and the
three numerical memory systems each append their own contract to the common
adapter-configuration instruction below. In the verbatim prompts that follow,
\emph{operator} is the implementation-level term for the selected memory system;
we preserve the wording used during evaluation.

\subsubsection{Common Adapter Configuration Instruction}

\begin{lstlisting}[style=prompt]
Configure the selected generic memory operator from visible task syntax.
Configure only choices admitted by the response schema. Numerical adapters use
the runtime's shared structured-output LLM extractor. Runtime code supplies
fixed version tags, operator identity, and extractor settings. Do not generate
parser code, task answers, hidden fields, rewards, or new operators.
\end{lstlisting}

\subsubsection{ReasoningBank request-boundary contract}

\begin{lstlisting}[style=prompt]
Operator: reasoning_bank_v1
Input contract: The current visible request and completed actor-visible
trajectories. Outcome labels are inferred by the actor model without access
to hidden task outcomes or external evaluator outputs.
Adapter configuration: Return only a syntax binding for isolating the changing
request from each full initial prompt. Set request_marker to a literal delimiter
that appears exactly once in every shown initial prompt, immediately before the
changing request. Prefer the closest stable delimiter to the request. Set it to
null only when the full prompt is itself the request or no shared delimiter is
available. Do not encode request content, task facts, tool names, answers, or
benchmark-specific rules.
\end{lstlisting}

\subsubsection{Keyed scalar regression contract}

\begin{lstlisting}[style=prompt]
Operator: hierarchical_regression_v1
Input contract: Scalar numeric outcomes with keyed dimensions, observation
times, requested prediction times, and provenance.
Adapter configuration: Use the shared llm_structured_output_v1 extractor to emit the
current time, every requested keyed prediction point, and only explicitly
visible realized outcomes. The first valid extraction establishes stable
semantic dimension names for the stream. Classify the target family as real,
positive_continuous, count, or probability, and choose whether its time effect
is absent or linear. Do not supply numerical priors, trends, noise levels, or
shrinkage constants. The runtime derives the link and output constraints,
estimates any global trend from repeated visible outcomes, and uses one
standardized regularization policy. Omit sources so runtime lifecycle defaults
apply.
\end{lstlisting}

\subsubsection{Discrete-event contract}

\begin{lstlisting}[style=prompt]
Operator: hierarchical_discrete_event_v1
Input contract: Monotone probability curves with categorical groups, features,
entity keys, weights, and provenance.
Adapter configuration: Return an llm_discrete_event_v1 manifest. Configure
interval_count and the names of stable categorical context roles. A context role
must describe a source-wide nuisance condition that repeats across multiple
measured entities; do not use entity identifiers, requested target features,
measurement availability, or one-off labels as context roles. The runtime
estimates its population baseline from visible evidence and applies one shared
weakly informative prior to every effect family. Do not supply numerical priors.
The shared runtime extractor derives target IDs and typed evidence records
directly from visible events, validates exact numeric provenance, derives a
compact response binding from each visible schema, and supplies generic
evidence-acquisition guidance. Do not generate parser code or encode concrete
target identifiers, context values, observations, answers, or rewards in the
manifest.
\end{lstlisting}

\subsubsection{Persistent-entity contract}

\begin{lstlisting}[style=prompt]
Operator: persistent_entity_state_v1
Input contract: Raw numeric identity coordinates, natural scale references,
numeric values, optional visible bounds, step index, and provenance.
Adapter configuration: Return an entity_regression_v1 manifest using the shared
structured-output extractor. Choose raw numeric fields that identify the same
latent entity across steps separately from numeric fields whose values should be
learned. Identity must contain the stable structural coordinates needed to
distinguish entities, including a size, span, or shape measurement when that
property defines the entity's persistent footprint. Put genuinely fluctuating
measurements in values instead. Do not omit a structural identity coordinate
merely because it also appears in the requested output. For each identity
coordinate, name a positive visible measurement that defines its natural scale,
such as an uncertainty, span, or size field, and declare whether that measurement
is already a standard deviation; a full uniform extent for the identity value
itself; or a full interval width whose midpoint locates a separate identity
coordinate. The operator converts that declared interpretation into unitless
uncertainty and learns residual calibration online; do not supply tolerances,
noise levels, match thresholds, or support thresholds. Classify predicted values
as real, positive_continuous, count, or probability; the runtime derives the link
and always regularizes temporal effects. Optional window bounds may filter
remembered entities to the currently visible numeric range. Give concise display
names for every identity and predicted value; the runtime owns the advisory text
and formatting. Omit sources because the runtime fixes ingestion at query start.
Do not use task names, hidden identities, rewards, or benchmark-tuned answers.
\end{lstlisting}

For numerical memory systems, the configuring LLM receives the following user
message. Terminal submissions are omitted because their values are candidate
answers rather than observations.

\begin{lstlisting}[style=prompt]
Selected operator: <operator identifier>

Visible prefix:
VISIBLE COMPLETED TRAJECTORIES

TRAJECTORY <index>

TURN <index> QUERY
<visible query>
ACTION
<visible action, or [terminal structured submission omitted]>
OBSERVATION
<visible observation>

...

CURRENT VISIBLE QUERY
<current instance prompt>
CURRENT RESPONSE SCHEMA
<current structured-response schema>
\end{lstlisting}

ReasoningBank needs only the changing request boundary, so its configuring LLM
instead receives the initial prompts from the completed and current instances:

\begin{lstlisting}[style=prompt]
Selected operator: reasoning_bank_v1

Visible prefix:
VISIBLE INITIAL PROMPTS

COMPLETED INITIAL PROMPT <index>
<completed initial prompt>

...

CURRENT INITIAL PROMPT
<current initial prompt>
\end{lstlisting}

If deterministic validation rejects an adapter configuration, one repair attempt adds:

\begin{lstlisting}[style=prompt]
The previous adapter configuration failed deterministic validation with: <validation error>
Return a corrected complete adapter configuration. Previous payload:
<previous structured payload>
\end{lstlisting}

\subsection{Typed Evidence Extractors}
\label{app:extractor-prompts}

After adapter configuration, the numerical operators use structured-output extractors to
translate visible text into operator-native records. The output schemas are,
respectively, \texttt{PanelTaskBatch},
\texttt{DiscreteEventEvidenceBatch}, and
\texttt{DetectionEvidenceBatch}. Deterministic validators admit only values
grounded in the supplied visible event.

\subsubsection{Keyed scalar regression extractor}

\begin{lstlisting}[style=prompt]
You are a memory extractor, not the task-solving actor. Translate one visible
numeric forecasting query into the complete typed task required by an
already-selected keyed regression operator. Return only the structured response.
current_time is the time explicitly identified as current. requested groups each
entity once with every target time the task asks the actor to predict. records
contains only historical outcomes explicitly revealed as feedback. Never treat
requested values, prior predictions, errors, scores, or tool-readable data as
realized observations. Choose short semantic dimension names and use the same
names exactly once in every requested point and observation. Copy all categorical
and numeric values exactly. Return an empty records list when no realized outcome
is visible. The framework validates structure and visible-value grounding before
anything enters memory.
\end{lstlisting}

\begin{lstlisting}[style=prompt]
Selected operator input contract:
- current_time: one visible numeric time
- requested: one series per entity with all requested target times
- records: only visible realized outcomes
- frozen dimension names: <existing names, or (infer)>

Translate this exact visible query:
<current visible query>
\end{lstlisting}

\subsubsection{Discrete-event extractor}

\begin{lstlisting}[style=prompt]
You are a memory extractor, not the task-solving actor. Convert exactly one
visible trajectory event into typed evidence for the already-selected
hierarchical discrete-event operator. Return only the structured response. Do
not write parser code, predict task answers, use rewards, or infer hidden values.
Infer response_binding directly from the visible response fields and schema
description: delimiter is the exact separator between an arbitrary target prefix
and component suffix, and component_labels must follow the semantic order stated
by the visible schema rather than property order. The compact rule must cover
every visible structured field. For every exact target identifier visibly
requested by the response schema or emitted in a completed structured action,
emit one target mapping. Its features are the smallest reusable conditions
visibly encoded as exact substrings of that identifier. Split combinations into
atomic conditions; do not include horizons, source labels, identifiers, or
inferred synonyms. Each record must cite the explicitly labeled zero-based
SOURCE TURN containing its numeric evidence; ignore any human-facing step number
inside the query. Copy an exact contiguous source_excerpt that names the
measurement and contains the probability vector and support weight. Copy the
observed monotone probability vector and weight exactly from that excerpt.
Observed zero and one values are valid: never replace them with epsilon, clipping,
or a probability floor. When both direct measurements and fitted, predicted, or
diagnostic values appear, record only direct measurements. Never use loss,
confidence, or diagnostic values as probabilities. Use entity_key for a direct
measurement of one requested target. Use raw_features only for conditions
declared in the target mappings. A record with no entity or features is a
baseline only when its excerpt explicitly identifies the complete source
population or global aggregate. Attach every configured categorical context role
to every record using literal values visible in the event. Emit a minimal
sufficient evidence set: at most one global baseline, direct measurements of
requested targets, and non-redundant reusable feature measurements. Skip
complements, residual buckets, alternative partitions, fitted proxies, and
ambiguous rows unless they directly measure a requested target or declared
feature. The framework derives response bindings separately.
\end{lstlisting}

\begin{lstlisting}[style=prompt]
Selected operator input contract:
- ordered probability dimensions: <interval count>
- infer a compact response binding from the visible response fields: emit the
  exact delimiter and component suffixes in their declared semantic order; the
  rule must apply to arbitrary target prefixes
- required categorical context roles: <configured roles>
- known target mappings from earlier visible events: <known mappings>
- acquisition guidance:
<actor-facing acquisition guidance>

Translate this exact visible event into evidence records:
EVENT KIND: instance_completed

STREAM INDEX: <cluster-local trajectory index>

CURRENT QUERY PROMPT:

CURRENT RESPONSE FIELDS:
[]

CURRENT RESPONSE SCHEMA:
{}

SOURCE TURN <zero-based index>

QUERY:
<visible query>

ACTION:
<visible JSON action, or STRUCTURED RESPONSE FIELDS followed by the JSON list of
terminal response field names>

OBSERVATION:
<visible observation>
\end{lstlisting}

\subsubsection{Persistent-entity extractor}

\begin{lstlisting}[style=prompt]
You are a memory extractor, not the task-solving actor. Convert one visible query
into typed measurement records for an already-selected multi-target tracking
operator. Return only the structured response. Copy the visible event identifier,
configured numeric context fields, record identifiers, and configured numeric
measurement fields exactly. Follow the configured semantic role names:
distinguish coordinates used for identity, values to predict, and any visible
numeric bounds or other context. Do not associate records across events, predict
persistent entities, transform units, or infer hidden values. Use every
configured field name exactly once and emit every visible measurement record.
The framework validates every value against the visible query.
\end{lstlisting}

\begin{lstlisting}[style=prompt]
Selected operator input contract:
- visible_id: the exact identifier of this measurement event
- numeric context roles: <configured context names>
- record_id: the exact visible identifier of each detection
- numeric measurement roles: <configured measurement names>
Map source field names into these semantic roles.

Translate this exact visible query into measurement records:
<current visible query>
\end{lstlisting}

When a structured evidence batch fails fixed schema or grounding validation, the
extractor appends the following repair text and retries within its fixed budget:

\begin{lstlisting}[style=prompt]
The previous evidence batch failed fixed schema or grounding validation. Return
a corrected batch.
Validation error: <exception type>: <validation error>
Rejected batch:
<previous structured batch, when available>
\end{lstlisting}

\subsection{ReasoningBank Memory Formation}
\label{app:reasoning-bank-prompts}

After each completed visible trajectory, ReasoningBank first asks the evaluated
model to infer whether the attempt succeeded or failed. This proxy judgment uses
only the visible request and trajectory, not hidden rewards or evaluator outputs.

\enlargethispage{\baselineskip}
\subsubsection{Self-judgment}

\begin{lstlisting}[style=prompt]
You are a helpful assistant that judges whether an agent successfully
completed a task. Use only the supplied task and trajectory. Return the
outcome and a concise rationale.
\end{lstlisting}

\begin{samepage}
Its user message is:

\begin{lstlisting}[style=prompt]
Task: <visible request>

Trajectory:
<completed actor-visible trajectory>

Did the agent successfully complete the task?
\end{lstlisting}
\end{samepage}

The response is a structured object containing \texttt{outcome}, either
\texttt{success} or \texttt{failure}, and a concise \texttt{rationale}.

\subsubsection{Memory extraction}

Given the proxy outcome and rationale, the same evaluated model receives one of
two fixed system prompts. For a successful trajectory, the system prompt is:

\begin{lstlisting}[style=prompt]
You are an expert in solving interactive tasks. You will be given a user query
and the corresponding trajectory that represents how an agent successfully
accomplished the task.

## Guidelines
Extract and summarize useful insights as memory items from the trajectory.
The goal is for the memory items to be helpful and generalizable for future
similar tasks.

## Important notes
- You must first think why the trajectory is successful, and then summarize
  the insights.
- Extract at most 3 memory items.
- Do not repeat similar or overlapping items.
- Prefer concrete, actionable procedures over abstract principles.
- Do not embed task-specific names, identifiers, queries, or literal values.

Return each item using the title, description, and content fields required by
the response schema. The description states when or when not to use the item.
The content contains one to three sentences of actionable insight: the insights
learned to successfully accomplish similar tasks in the future.
\end{lstlisting}

For a failed trajectory, the system prompt is:

\begin{lstlisting}[style=prompt]
You are an expert in solving interactive tasks. You will be given a user query
and the corresponding trajectory that represents how an agent attempted to
resolve the task but failed.

## Guidelines
Extract and summarize useful insights as memory items from the trajectory.
The goal is for the memory items to be helpful and generalizable for future
similar tasks.

## Important notes
- You must first reflect and think why the trajectory failed, and then summarize
  what lessons you have learned or strategies to prevent the failure in the
  future.
- Extract at most 3 memory items.
- Do not repeat similar or overlapping items.
- Prefer concrete, actionable procedures over abstract principles.
- Do not embed task-specific names, identifiers, queries, or literal values.

Return each item using the title, description, and content fields required by
the response schema. The description states when or when not to use the item.
The content contains one to three sentences of actionable insight: the insights
learned to avoid such failures and successfully accomplish similar tasks in the
future.
\end{lstlisting}

Both prompts receive the following user message:

\begin{lstlisting}[style=prompt]
**Query:** <visible request>

**Trajectory:**
<completed actor-visible trajectory>

The task <succeeded or failed> because: <self-judgment rationale>
\end{lstlisting}

The structured response contains at most three items, each with
\texttt{title}, \texttt{description}, and \texttt{content} fields.

\subsection{Actor-Facing Context Templates}
\label{app:actor-context-templates}

Context Memory Adapter injects the generated advisory into the actor's context while
leaving the native instance prompt unchanged. The generated rows depend on
visible state; the fixed wrapper text is reproduced below.

\subsubsection{ReasoningBank}

\begin{lstlisting}[style=prompt]
Below are some memory items accumulated from past interaction with the
environment that may help solve the current task. Use them only when relevant.

# Memory Item 1
## Title <title>
## Description <description>
## Content <content>
\end{lstlisting}

ReasoningBank retrieves one prior experience and injects all memory items
extracted from it.

\subsubsection{Keyed scalar regression}

\begin{lstlisting}[style=prompt]
=== PYMC HIERARCHICAL PREDICTIVE STATE ===
This advisory state uses only canonical previous-round outcomes visible in task
prompts and prior structured submissions. Inspect the current files and revise
these estimates when warranted.
Entity forecasts from one shared global/entity/factor/trend rule:
- <entity dimensions>: visible outcomes {<recent outcomes>}; advisory forecasts
  {<time=value, uncertainty, and support>}
Resolved prediction receipts: n=<count>, mean(actual-predicted)=<mean residual>,
mean absolute residual=<mean absolute residual>.
Memory estimates are not ground truth and are not a submission.
=== END PYMC HIERARCHICAL PREDICTIVE STATE ===
\end{lstlisting}

The resolved-prediction line is omitted when no prior advisory prediction has a
visible outcome.

\subsubsection{Discrete event}

\begin{lstlisting}[style=prompt]
=== HIERARCHICAL EVENT MEMORY ===
Use only visible source data. Obtain a source-wide event-free curve and useful
entity marginals or combinations when tools permit. Preserve exact support
counts and canonical conditions from requested identifiers. Current-source
aggregates may reflect selection effects rather than population truth; the final
estimates pool evidence across sources.
=== END HIERARCHICAL EVENT MEMORY ===

Prior model estimates by target:
- <target identifier>: [<ordered event-free probabilities>]
\end{lstlisting}

Before the model has fitted state, the final block is replaced by
\texttt{No prior model state is available yet.}

\subsubsection{Persistent entity}

\begin{lstlisting}[style=prompt]
=== ONLINE ENTITY STATE ===
Candidate persistent-entity hypotheses inferred from prior visible measurements
and updated with the current measurement set are listed below. Candidates are
uncertain evidence and may represent duplicates or noise. Reconcile compatible
candidates and produce the required structured response yourself.
- candidate=<entity>; <identity coordinate and observed range>; support=<count>;
  seen_rounds=<range>; active_now=<boolean>; <predicted values>
=== END ONLINE ENTITY STATE ===
\end{lstlisting}

\end{document}